\documentclass[runningheads]{llncs}

\usepackage{graphicx}
\usepackage{cite}
\usepackage{multirow}
\usepackage{longtable}
\usepackage{float}
\usepackage[utf8]{inputenc} 
\usepackage[T1]{fontenc}    
\usepackage{hyperref}       
\usepackage{cleveref}
\usepackage{url}            
\usepackage{booktabs}       
\usepackage{amsfonts}       
\usepackage{nicefrac}       
\usepackage{microtype}      
\hypersetup{colorlinks,urlcolor=blue}
\begin{document}
\title{Difficulty Translation in Histopathology Images}

\titlerunning{Difficulty Translation in Histopathology Images}
%
\author{Jerry Wei\inst{1}
\and Arief Suriawinata\inst{2}
\and Xiaoying Liu\inst{2}
\and Bing Ren\inst{2}
\and Mustafa Nasir-Moin\inst{1}
\and Naofumi Tomita\inst{1}
\and Jason Wei\inst{1}
\and Saeed Hassanpour$^\dagger$\inst{1}
}
\authorrunning{Wei et al.}
%
\institute{\hspace{-1.5mm}Dartmouth College, Hanover, NH 03755, USA \and
\hspace{-1mm}Dartmouth-Hitchcock Medical Center, Lebanon, NH 03756, USA
\email{$^\dagger$saeed.hassanpour@dartmouth.edu}
}
\maketitle              
\begin{abstract}
The unique nature of histopathology images opens the door to domain-specific formulations of image translation models.
We propose a difficulty translation model that modifies colorectal histopathology images to be more challenging to classify.
Our model comprises a scorer, which provides an output confidence to measure the difficulty of images, and an image translator, which learns to translate images from easy-to-classify to hard-to-classify using a training set defined by the scorer.
We present three findings. 
First, generated images were indeed harder to classify for both human pathologists and machine learning classifiers than their corresponding source images. 
Second, image classifiers trained with generated images as augmented data performed better on both easy and hard images from an independent test set.
Finally, human annotator agreement and our model's measure of difficulty correlated strongly, implying that for future work requiring human annotator agreement, the confidence score of a machine learning classifier could be used as a proxy.

\end{abstract}

\section{Introduction}
Automated histopathology image analysis has advanced quickly in recent years \cite{Coudray2017,Ehteshami2017,Tomita2019,WeiJason2019} due to substantial developments in the broader fields of deep learning and computer vision \cite{He2015,Krizhevsky2012,Szegedy2014}.
While histopathology imaging research typically applies these general computer vision models directly and without modification, there may be domain-specific models that might not generalize to broader computer vision tasks but can be useful for specifically analyzing histopathology images. 

In this study, we formulate a difficulty translation model for histopathology images, i.e., given a histopathology image, we aim to modify it into a new image that is harder to classify.
Our model is motivated by the observation that histopathology images exhibit a range of histological features that determines their histopathological label. 
For instance, both an image with small amounts of sessile serrated architectures and an image covered by sessile serrated architectures would be classified by a pathologist as a sessile serrated adenoma.
We know that this range of features exists because normal tissue progressively develops precancerous or cancerous features over time, which differs from general domain datasets such as ImageNet \cite{Deng2009}, in which classes are distinct by definition (there is no range of cats and dogs, for instance). 
This continuous spectrum of features allows us to use the confidence of a machine learning classifier to determine the amount and intensity of cancerous features in an image. 
In other words, the confidence of the classifier can act as a proxy for the extent of histological features in an image).
In this paper, we propose and evaluate a difficulty translation model that generates hard-to-classify images that are useful as augmented data, thereby demonstrating a new way to exploit the unique nature of histopathology images. 
We post our code publicly at \url{https://github.com/BMIRDS/DifficultyTranslation}.

\begin{figure}[t!]
\centering
    \setlength{\belowcaptionskip}{-10pt}
    \setlength{\abovecaptionskip}{4pt}
    \includegraphics[width=\linewidth]{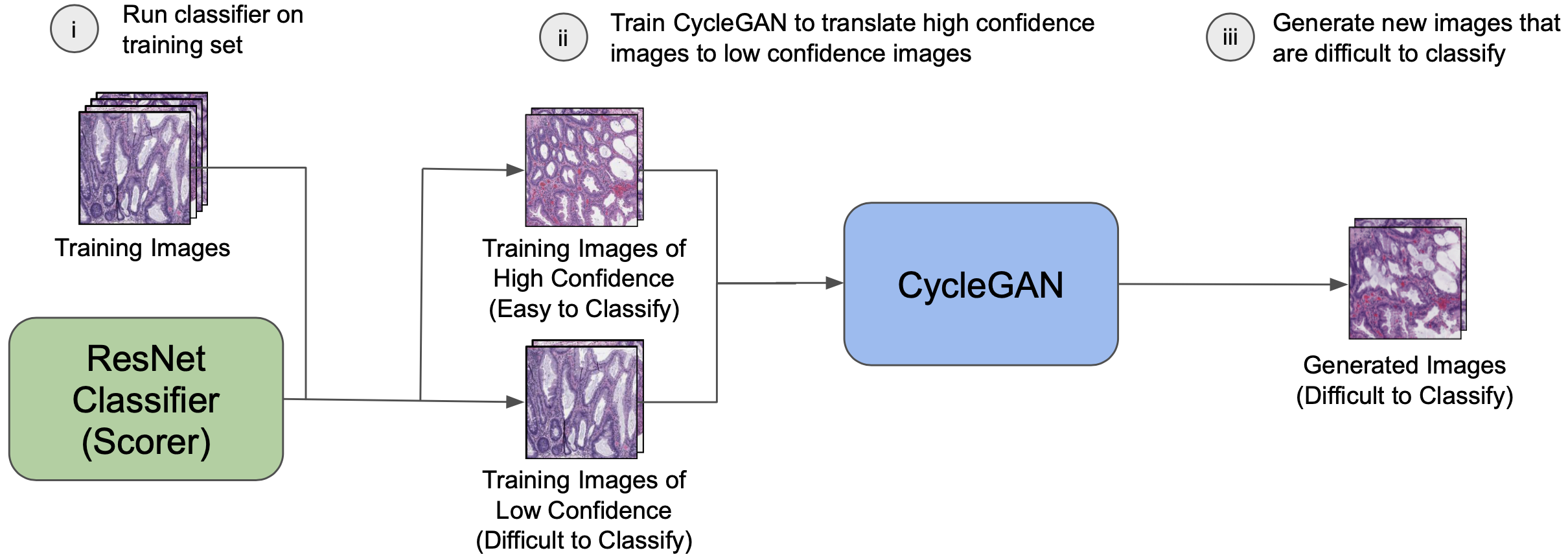}
    \caption{Our proposed model for modifying colorectal histopathology images to be more challenging to classify.
    }
    \label{fig:methodoverview}
\end{figure}

\section{Methods and Materials}
\textbf{Problem set-up and model.}
Given some training image $x_i$ of class $X$, we aim to generate $\tilde x_i$, which maintains the same histopathological class and general structure of $x_i$, but is more challenging to classify. 
We propose a model that comprises two networks: a scorer, which predicts the difficulty $c(x_i)$ of some image $x_i$ \cite{Hacohen2019,Weinshall2018}, and an image translator, which translates images that are easy to classify into images that are harder to classify. 
In this study, we use ResNet-18 \cite{He2015} as the scorer for colorectal histopathology images and train it to convergence on the downstream task of hyperplastic polyp/sessile serrated adenoma classification, and we assign $c(x_i)$ as the softmax output (confidence) of class $X$. 
For the image translator, we use a cycle-consistent generative adversarial network (CycleGAN), which learns the mapping $G: \hat X \rightarrow \tilde X$, where we assign $\hat X$ as the class for the set of images $\{\hat x\}$ such that $c(\hat x_i)$ is high for all $\hat x_i \in \{\hat x\}$ (easy-to-classify images) and $\tilde X$ as the class for the set of images $\{\tilde x\}$ where $c(\tilde x_j)$ is low for all $\tilde x_j \in \{\tilde x\}$ (hard-to-classify images). 
With this configuration, given some image $\hat x_i \in \{\hat x\}$, we can generate a similar example $\tilde x_i$ that maintains the same histopathological class but is harder to classify.
An overview schematic of our model is shown in Figure \ref{fig:methodoverview}.
Hereafter, we use the terms \textit{easy images} and \textit{hard images} to refer to images that are easy to classify and hard to classify based on a pre-trained classifier's confidence output.\footnote{In this metric for measuring the difficulty of images, the pre-trained classifier does not classify an image as easy or hard, but rather classifies an image as HP or SSA---the classifier's \textit{confidence} on its HP or SSA prediction determines whether the image is considered to be easy or hard to classify.}
When referring to easy and hard as perceived by annotators, we use the terms \textit{high-agreement} images and \textit{low-agreement images}, which represent 3/3 annotator agreement and 2/3 annotator agreement, respectively.

\textbf{Dataset.} 
For our experiments, we first collected and scanned 328 Formalin-Fixed Paraffin-Embedded (FFPE) whole-slide images of colorectal polyps, originally diagnosed as either hyperplastic polyps (HPs) or sessile serrated adenomas (SSAs), from patients at the Dartmouth-Hitchcock Medical Center, our tertiary medical institution. 
From these 328 whole-slide images, we then extracted 3,152 patches (portions of size $224\times224$ pixels from whole-slide images) representing diagnostically relevant regions of interest for HPs or SSAs. 
Three board-certified practicing gastrointestinal pathologists at the Dartmouth-Hitchcock Medical Center independently labeled each image as HP or SSA. The use of the dataset in this study was approved by our Institutional Review Board (IRB).

The gold standard label for each image was determined by the majority vote of the labels from three pathologists.
Table \ref{tab:dataset} shows the distribution of high-agreement and low-agreement images for each class in the training and test set. 
Note that our dataset is imbalanced because SSAs naturally occur less frequently than HPs.
Figure \ref{fig:datasetexamples} shows examples of high-agreement and low-agreement images from each class.
Images were split randomly by whole slide into the training set and test set, so images from the same whole slide either all went into the training set or all went into the test set.
We chose the task of sessile serrated adenoma detection, which is challenging and clinically important for colonoscopy, one of the most common screening tests for colorectal cancer \cite{Rex2017}. 
We used a training set of 2,051 images and a test set of 1,101 images, of which each image was labeled as either hyperplastic polyp (HP) or sessile serrated adenoma (SSA). 

\begin{table}[t!]
\setlength{\tabcolsep}{0.5em}
\setlength{\abovecaptionskip}{5pt}
\centering
\caption{Distribution of data in our training and test sets based on the level of annotation agreement among three pathologist annotators. HP: hyperplastic polyp, SSA: sessile serrated adenoma.}
\begin{tabular}{l| c c c |c c c}
    \toprule
    & \multicolumn{3}{|c|}{Training Set Images} & \multicolumn{3}{|c}{Test Set Images} \vspace{0.04 in} \\
    Level of Agreement & HP & SSA & Total & HP & SSA & Total \\
    \midrule
    2/3 Annotators & 670 & 173 & 843  & 316 & 89 & 405 \\
    3/3 Annotators & 860 & 348 & 1,208 & 492& 204 & 696 \\
    \midrule
    Total & 1,530 & 521 & 2,051 & 808 & 293 & 1,101 \\
    \bottomrule
\end{tabular}
\label{tab:dataset}
\vspace{-2mm}
\end{table}

\begin{figure}[ht]
\centering
    \setlength{\abovecaptionskip}{5pt}
    \includegraphics[width=0.7\linewidth]{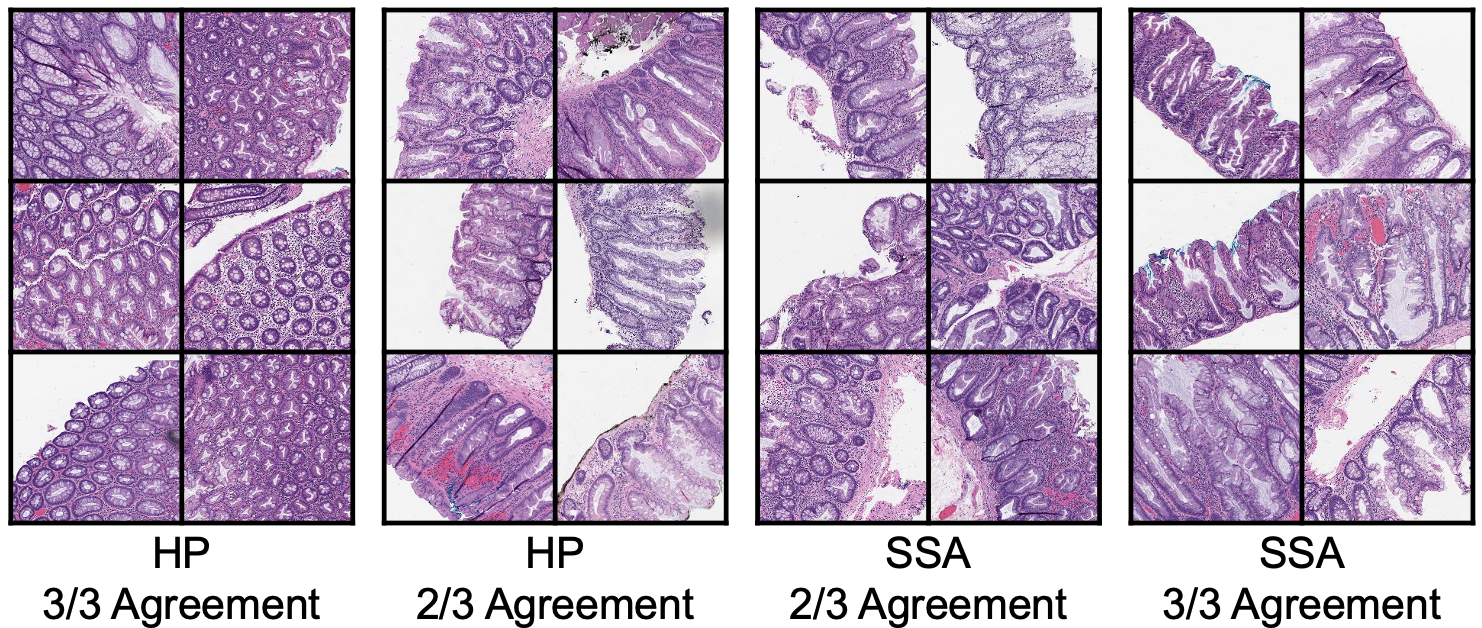}
    \caption{Examples of high-agreement and low-agreement images for the hyperplastic  (HP) and sessile serrated adenoma (SSA) classes.}
    \label{fig:datasetexamples}
\end{figure}

\begin{figure}[ht]
\centering
    \setlength{\belowcaptionskip}{-10pt}
    \setlength{\abovecaptionskip}{4pt}
    \includegraphics[width=0.7\linewidth]{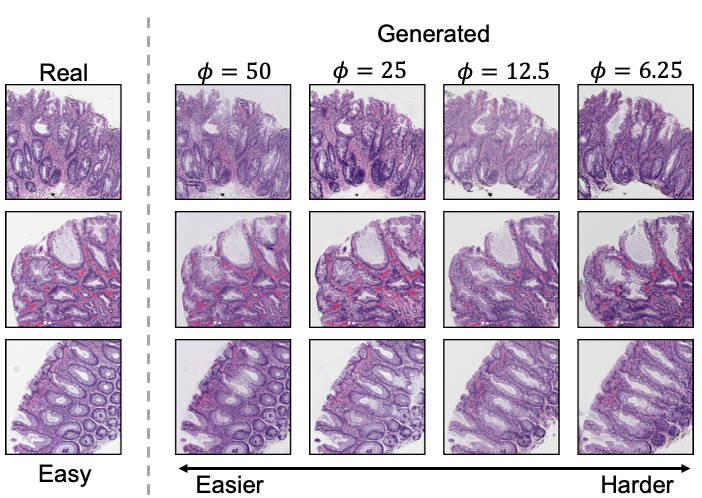}
    \caption{Generated hyperplastic polyp (HP) images that are intended to be more difficult to classify than their real source images. 
    $\phi$ is the selectivity parameter for target domain images used to train our image translator---images generated using a lower $\phi$ parameter are intended to be more difficult to classify. 
    }
    \label{fig:visualablation}
\end{figure}

\section{Experiments}
\subsection{Generating more challenging training examples}
In our image translation model, we define a selectivity parameter, $\phi$, as the percent of training set images used as hard data in the target domain for training our image translator.
For instance, at $\phi$ = 50, the lower 50$\%$ of training set images by confidence, as determined by our pre-trained classifier, was used as target domain training data for the image translator.
We train our image translation model for various $\phi$ to generate difficult HP images.
We find that for a given image classifier, generated images---particularly those with lower $\phi$---were indeed harder than their corresponding source images to classify (Figure \ref{fig:histograms} in the Appendix).
Figure \ref{fig:visualablation} shows examples of images that were generated with varying $\phi$.
While images generated with lower $\phi$ are typically harder to classify, at very low $\phi$, generated images are no longer representative of their target class (e.g., generated HP images begin to look like SSA images).
We therefore recommend using the smallest $\phi$ possible such that the original class label is generally maintained and a sufficient number of images is provided to train the image translation model.
For our dataset, we find that the model needs to be trained with at least 100 images $(\phi>6.25)$.

We also evaluated the difficulty of generated images by presenting them to three board-certified gastrointestinal pathologists for manual evaluation in a blinded test. 
Using labels where the top 50\% and bottom 50\% of images marked as HP by confidence of our pre-trained classifier were considered to be easy HP images and hard HP images, respectively, we randomly sampled 75 easy HP images, 75 hard HP images, 75 generated HP images that were translated from the selected easy HP images, and 75 SSA images from our training set.
Each pathologist then independently classified each image as HP or SSA.
As shown in Table \ref{tab:pathtest}, pathologists disagreed more (2/3 annotator agreement) on generated images (22.7$\%$) than their real counterparts (13.3$\%$), although not as much as they did for real hard images (26.7$\%$).
At the same time, generated images retained their original class label of HP $96\%$ of the time based on annotator agreement.

\begin{figure}[h!]
    \centering
    \includegraphics[width=0.56\linewidth]{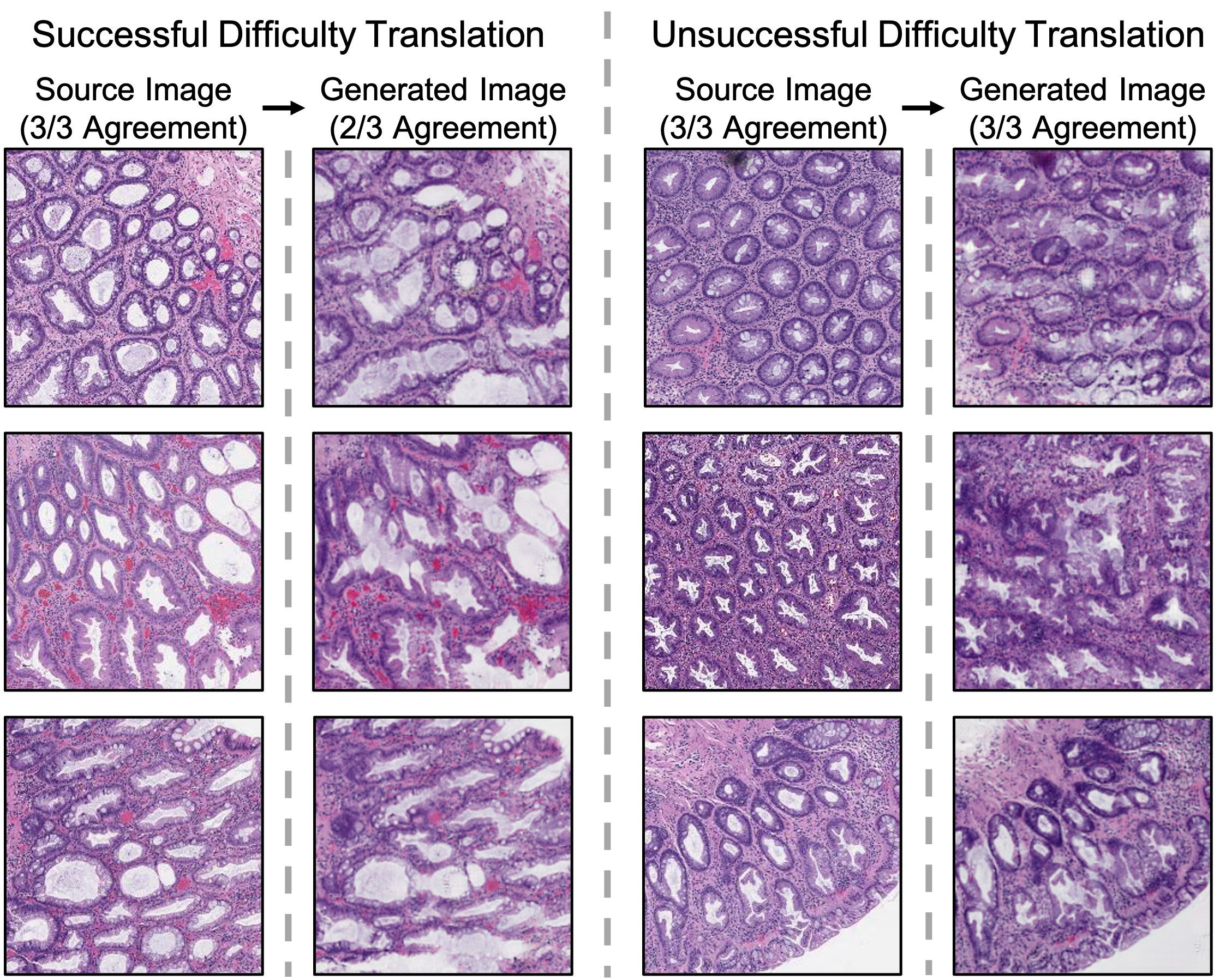}
    \caption{Generated images, which were intended to be more difficult to classify, had lower inter-annotator agreement between pathologists. The left set of images shows translations that successfully lowered inter-annotator agreement, while the right set of images shows translations that did not lower the inter-annotator agreement. In all examples, generated images retained their original ground-truth labels.} 
    \label{fig:testexamples} 
\end{figure}

To make Table \ref{tab:pathtest} more readable, we omit the classification results for SSA images. 
The proportion of the 75 images in our blinded test with an original ground truth of SSA that were again marked as SSA by a majority of pathologists during the blinded test was 89.3$\%$, indicating that our pathologist annotators were relatively consistent in their classifications.
In Figure \ref{fig:testexamples}, we show examples of translations that were successful and unsuccessful in making images more difficult to classify.

\begin{table}[t!]
\setlength{\tabcolsep}{0.5em}
\setlength{\abovecaptionskip}{5pt}
\centering
\caption{Pathologists agreed less on ground-truth labels for generated HP images compared with their real image counterparts. Most generated images maintained their HP label in that both source images and their corresponding generated images were classified as HP by the majority of annotators.}
\begin{tabular}{l c c | c }
    \toprule
    & \multicolumn{2}{c}{Pathologist Annotator Agreement (\%) \vspace{0.06 in}} & \\
    Image Type & \nicefrac{2}{3} Agreement & \nicefrac{3}{3} Agreement & Maintained Label (\%)\\
     \midrule
    Real--Easy & 13.3 & 86.7 & 100.0\\
     Real--Hard & 26.7 & 64.0 & 90.7\\
     \midrule
     Generated--Hard & 22.7 & 73.3 & 96.0\\
    \bottomrule
\end{tabular}
\label{tab:pathtest}
\end{table}
\raggedbottom

{}

\subsection{Improving the Performance of Classifiers} 
We conduct further experiments to explore using the generated images as additional data to better train a classifier.
Given a training set, we use the easy HP images as source images to generate harder HP images. 
Of these generated harder images, we use the images that maintain the HP class label, according to our pre-trained classifier, as additional data for training a new classifier.

For all image classifiers, we train ResNet-18 \cite{He2015} for 50 epochs (far past convergence) using the Adam optimizer \cite{Kingma2014} with a L2 regularization factor of $10^{-4}$.
We use an initial learning rate of $10^{-3}$, decaying by 0.91 every epoch.
Every trained model used automatic data augmentation of online color jittering uniformly sampled from the range of $\pm0.5$ for brightness, $\pm0.5$ for contrast, $\pm0.2$ for hue, and $\pm0.5$ for saturation, as implemented in PyTorch.

Table \ref{tab:improvement} shows the performance of classifiers trained with our generated images as additional data compared with the baseline of standard training on the original dataset as well as a naïve data augmentation technique of directly combining parts from easy and hard images \cite{Summers2019}. 
In order to account for variance in random weight initializations and performance fluctuations throughout training, we run each configuration for twenty random seeds, and for each seed we record the mean of the five highest AUC scores, which are calculated for every epoch.
Notably, adding images generated at $\phi=25$ consistently outperformed naïve data augmentation and no data augmentation for both low-agreement and high-agreement images.
We posit that naïve data augmentation was unsuccessful in this case because the features in the augmented data were not reflected in the test set. 

\begin{table}[t!]
\setlength{\tabcolsep}{0.5em}
\setlength{\abovecaptionskip}{5pt}
\centering
\caption{Performance (\% AUC $\pm$ standard error) of image classifiers trained with generated images as augmented data on test set images with high annotator agreement, low annotator agreement, and all test set images.}
\begin{tabular}{l c c c}
    \toprule& \multicolumn{3}{c}{Test Set Performance} \\
    Training dataset & High-Agreement & Low-Agreement & All Images \\
     \midrule
    Unmodified original dataset & 91.3 $\pm$ 0.2& 66.0 $\pm$ 0.6 & 83.1 $\pm$ 0.2\\ 
     + naïve data augmentation  & 90.9 $\pm$ 0.2 & 64.3 $\pm$ 0.5 & 82.1 $\pm$ 0.3\\
    \midrule
     + generated images, $\phi=50$ & 91.9 $\pm$ 0.2 & 66.6 $\pm$ 0.4 & 83.8 $\pm$ 0.3\\
     + generated images, $\phi=25$  & \textbf{92.6 $\pm$ 0.2} & \textbf{68.1 $\pm$ 0.7} & \textbf{84.8 $\pm$ 0.4}\\
     + generated images, $\phi=12.5$ & 91.7 $\pm$ 0.2 & 65.5 $\pm$ 0.4 & 83.4 $\pm$ 0.2\\
    \bottomrule
\end{tabular}
\label{tab:improvement}
\end{table}

\subsection{Comparing Machine and Human Difficulty Measures}
\begin{figure}[ht]
\centering
    \setlength{\belowcaptionskip}{-20pt}
    \setlength{\abovecaptionskip}{1pt}
    \includegraphics[width=\linewidth]{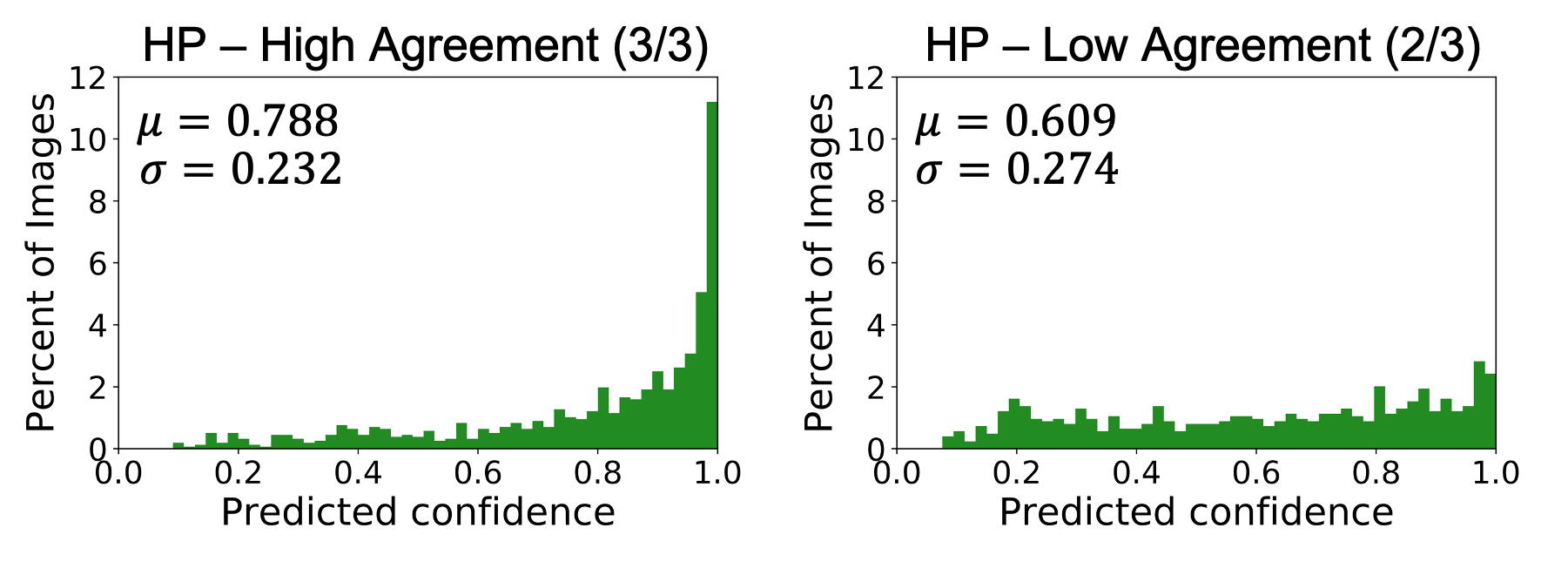}
    \caption{Distributions of predicted confidences of a pre-trained classifier vastly differed for hyperplastic polyp (HP) images with low (2/3) and high (3/3) annotator agreement.}
    \label{fig:easy_hard_hist}
    \vspace{0.1in}
\end{figure}

While it was previously unconfirmed whether the confidence output of a machine learning model correlates with the human concept of difficulty, for our dataset, we find that the confidence of our pre-trained classifier indeed correlates strongly with human annotator agreement. 
As shown in Figure \ref{fig:easy_hard_hist}, the predicted confidence distribution of images with high annotator agreement vastly differs from that of images with low annotator agreement.
We compared these distributions using a Kolmogorov-Smirnov test for equality of two distributions \cite{Massey1951} and computed a Kolmogorov-Smirnov statistic of 0.302 over all 1530 HP images in the training set with a statistically significant p-value of $p = 1.5 \times 10^{-30}$, indicating that the two distributions are not equal.
The correlation between these two measures of difficulty implies that for tasks requiring human annotator agreement data, the confidence of a machine learning classifier, which is computed automatically, could be used as a reliable proxy.

\section{Related Work and Discussion}
Generative adversarial networks (GANs) have been used in deep learning for medical imaging to generate synthetic data ranging from MRIs to CT scans \cite{Dar2018,Salehinejad2018,Wang2018}.
For histopathology, several studies used GANs for both image generation and translation \cite{Bayramoglu2018,Bentaieb2017,Burlingame2018,Cho2017,Jackson2020,Zanjani2018}.
Along the same lines as our work, GAN-generated images have been used as augmented data for liver lesion \cite{Frid-Adar2018}, bone lesion \cite{Gupta2019}, and rare skin condition classification \cite{Ghorbani2019}.
While prior work generates augmented data as a means to improve general performance, the augmented data that we generate aims to help classifiers specifically on examples that are challenging. 
Moreover, our methodology, to our knowledge, substantially differs from previous work due to its focus on example difficulty.

This paper advances related work from our group, which is focused on colorectal polyp classification \cite{Korbar2017, WeiJason2020}  and used image translation between different colorectal polyp types to address data imbalances \cite{Wei2019}. 
In this study, we translate images within the same class to become more difficult to differentiate from other classes, arguing that the range of features in histopathology images can and should be utilized to train better-performing machine learning models.

Of possible limitations, measuring whether generated images maintained the same quality and realistic features as real images is challenging.
Although generated images occasionally contained minor mosaic-like patterns, they remained readable and improved classifier training over baseline augmentation methods, suggesting that useful histologic features were retained.
Also, another approach for difficulty translation could be to directly use human annotator agreement to translate images from high to low agreement. 
In our paper, however, we define difficulty according to the confidence of a pre-trained classifier, since this framework generalizes to cases where annotator agreement data is unavailable.

To conclude, this work shows how to generate difficult yet meaningful training data by exploiting the range of features in histopathology images. 
Future research could explore difficulty translation in the context of curriculum learning \cite{Bengio2009} or defending against adversarial attacks \cite{Ma2019}.
This study and its results encourage further research to make use of the range of features in histopathology images in more creative ways.

\section{Acknowledgments}
This research was supported in part by the National Institute of Health grants (R01LM012837, R01CA098286, and P20GM104416).

\newpage
\section{Appendix}

\begin{figure}[H]
    \centering
    \setlength{\abovecaptionskip}{5pt}
    \includegraphics[width=\linewidth]{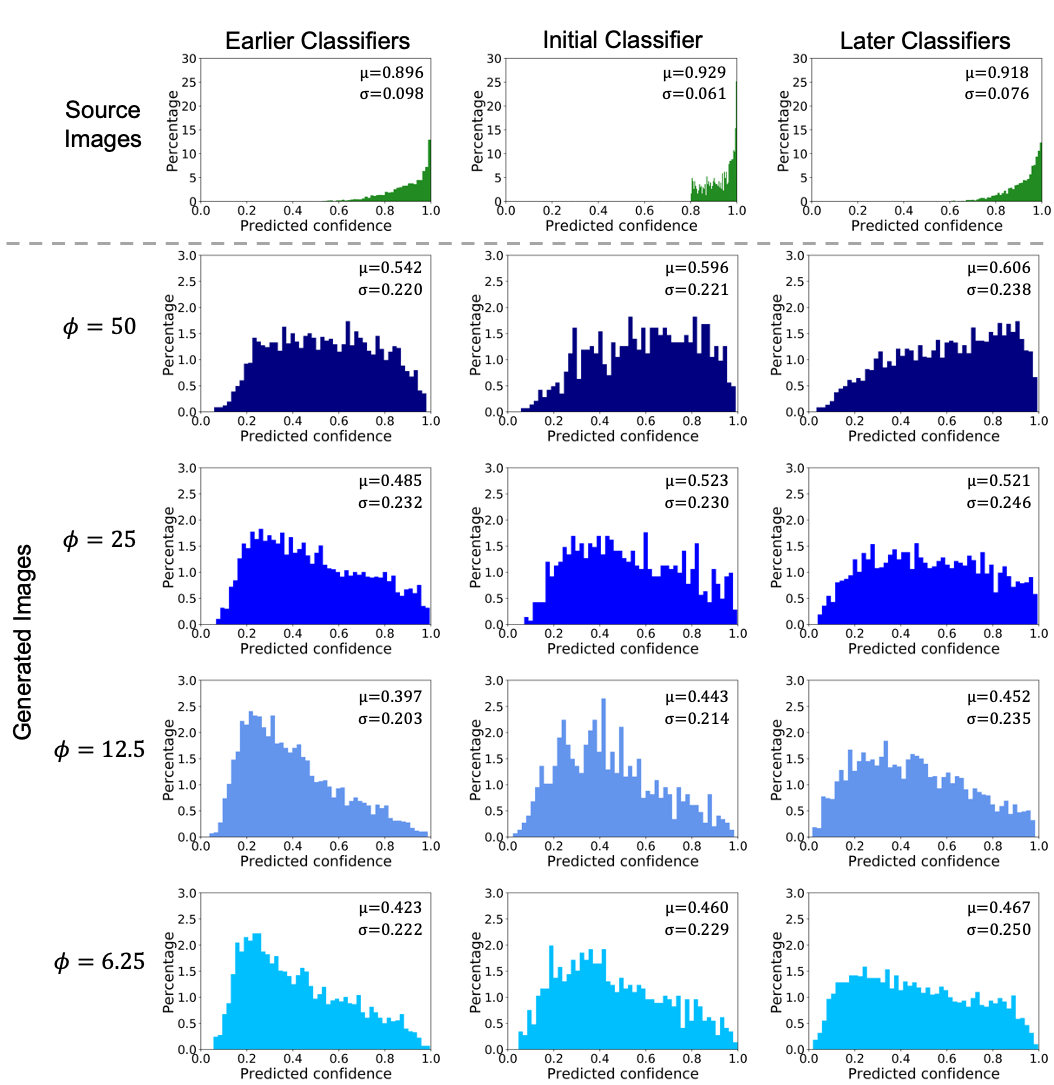}
    \caption{Image classifiers had less confident predictions on images generated by our image translation model than on real source hyperplastic (HP) images in the training set. $\phi$ is the selectivity parameter for target domain images used to train our image translator---images generated using a lower $\phi$ parameter are intended to be more difficult to classify. We show the predicted distribution of machine learning classifiers with varying amounts of training, since classifiers with more training tend to have more confident predictions. The initial classifier is the original classifier used to define the training set for our image translation model, which was trained for 25 epochs. Earlier classifiers is the average of classifiers trained for 17, 19, 21, and 23 epochs, and later classifiers is the average of classifiers trained for 27, 29, 31, and 33 epochs.}
    \label{fig:histograms}
\end{figure}

\newpage
\newpage
\bibliography{bibliography}
\bibliographystyle{splncs04}

\end{document}